\documentclass[letterpaper, 10 pt, conference]{config/ieeeconf}  % Comment this line out if you need a4paper

\IEEEoverridecommandlockouts                              % This command is only needed if 
\usepackage{etoolbox}
\makeatletter 
\let\NAT@parse\undefined
\makeatother
\usepackage[numbers]{natbib} 

\usepackage{float}

\usepackage{multirow}
\usepackage[bookmarks=true, colorlinks=false,
    linkcolor=black,
    filecolor=magenta,      
    urlcolor=cyan]{hyperref}
\usepackage{url}
\usepackage{graphicx} % For importing graphics 
\usepackage{amsfonts,amssymb,amsmath} 
\usepackage{graphics} 
\usepackage{epsfig} 
\usepackage[all]{xy}  
\usepackage{subcaption}
\usepackage{algorithm}
\usepackage{algpseudocode}

\usepackage[labelfont=small,textfont={small}]{caption}
\usepackage{subcaption}
\usepackage{epstopdf} 
\usepackage{balance}
\usepackage{soul}
\usepackage{color}
\usepackage{pgf}
\usepackage{adjustbox}
\usepackage{pgffor}
\usepackage{etoolbox}
\usepackage{tabularx}
\usepackage{gensymb}
\usepackage{pifont}

\usepackage{booktabs}
\usepackage{multirow, multicol}

\title{
Structural Damage Detection Using AI Super Resolution and Visual Language Model }
\author{Catherine Hoier and Khandaker Mamun Ahmed\\
The Beacom College of Computer and Cyber Sciences, Dakota State University\\
Madison, South Dakota, USA\\
Emails: \{Catherine.Hoier@trojans, khandakermamun.ahmed@\}dsu.edu
}

\begin{document}
\setlength{\textfloatsep}{-2pt}  % Space between the text and a figure at the top/bottom
\setlength{\floatsep}{-2pt}      % Space between figures (or tables) and the text
\maketitle
% \thispagestyle{empty}
% \pagestyle{empty}
% \maketitle              % typeset the title of the contribution

%POTENTIAL SUBMISSION - DUE END OF JULY - 5th Cyber Awareness and Research Symposium (CARS): Paper submission deadline: July 30, 2025; https://www.ieee\-cars.org/authors/call-for-papers https://www.ieee-cars.org/

%POTENTIAL SUBMISSION - DUE JULY 28TH AAAAAAA 
%https://icmla-conference.org/icmla25/keydates.html

%TODO: gpt for professionaliism and academic sounding - NOTE ASK DR. AHMED IF HE HAS DONE THIS, otherwise this will take like 5 minutes
%bar charts about data collected and results 
%DONE: show specific accuracy in abstract instead of saying "very good"
%publish script of automation of this process to github and put that in paper

%NOTE. 8 PAGE HARD MAX, INCLUDING CITATION PAGE. NEED TO CURRENT CUT OUT A FEW SENTENCES AS EVERYTHING IS CURRENTLY WRITTEN. 

\newcommand{\mamun}[1]{\textbf{{\color{red}Comment: #1}}}
\newcommand{\catherine}[1]{\textbf{{\color{green}catherine: #1}}}

\begin{abstract}
Natural disasters pose significant challenges to timely and accurate damage assessment due to their sudden onset and the extensive areas they affect. Traditional assessment methods are often labor-intensive, costly, and hazardous to personnel, making them impractical for rapid response, especially in resource-limited settings. This study proposes a novel, cost-effective framework that leverages aerial drone footage, an advanced AI-based video super-resolution model, Video Restoration Transformer (VRT), and Gemma3:27b, a 27 billion parameter Visual Language Model (VLM). This integrated system is designed to improve low-resolution disaster footage, identify structural damage, and classify buildings into four damage categories, ranging from no/slight damage to total destruction, along with associated risk levels. The methodology was validated using  pre- and post-event drone imagery from the 2023 Turkey earthquakes (courtesy of The Guardian) and satellite data from the 2013 Moore Tornado (xBD dataset). The framework achieved a classification accuracy of 84.5\%, demonstrating its ability to provide highly accurate results. Furthermore, the system's accessibility allows non-technical users to perform preliminary analyses, thereby improving the responsiveness and efficiency of disaster management efforts.
%demonstrating its ability to provide rapid, scalable, and relatively low-cost post-disaster assessments. Furthermore, the system's accessibility allows non-technical users to perform preliminary analyses, thereby improving the responsiveness and efficiency of disaster management efforts. %Our code is available at {\color{magenta}\url{https://github.com/}}
%TODO: GET THE SCRIPT AND PUT IT ON GITHUB ASAP. LIKE A MONTH AGO THIS SHOULD'VE BEEN DONE CAT. FIX THIS NOW!!!!!

%{\color{red} We mentioned this system provides rapid, scalable and low-cost damage assessment. How is this system rapid? There is no information of rapidness was given in the result section? How is it scalable? Same goes for low cost. We did not provide any cost information in the result section, I believe.}

%Though the computer components required for efficient processing of the images and running of the VLM as well as the drones to capture the footage can cost thousands of dollars, the amount of costs it saves on recovery workers and professional damage assessment professionals can save multiple times more then those initial costs.

\end{abstract}

\section{Introduction}
\vspace{-3pt}
Natural disasters such as earthquakes, tsunamis, and volcanic eruptions often occur without warning, leading to catastrophic consequences including significant infrastructure damage, injuries, and loss of life within a matter of minutes. In the immediate aftermath of such events, accurate and timely damage assessments are essential for evaluating the severity of the disaster, prioritizing emergency response efforts, and initiating long-term recovery plans to save lives and restore critical services as quickly as possible \cite{WorldBank_BuildingBackBetter_2018}. However, physical access to disaster-affected areas is often hampered by collapsed infrastructure, blocked roads, or other environmental obstructions \cite{mcentire2006managing}. Even when access is possible, dangerous conditions such as extreme heat, biohazards, and unstable structures pose significant safety risks to first responders \cite{doi:10.1177/216507990705500603}. Additionally, manual assessment procedures are often labor-intensive, time-consuming, and costly, delaying recovery efforts and stretching limited resources.

Aerial drones equipped with video recording capabilities offer a promising solution to these challenges. These drones can bypass ground-level obstacles and operate in hazardous environments, enabling rapid data collection over large areas without endangering human life. This approach not only enhances the efficiency of disaster response but also provides critical visual data that can be processed automatically for faster decision-making. Despite the well-known impact of natural disasters, there remains a concerning lack of standardized and scalable procedures for assessing damage, especially across a wide range of disaster scenarios \cite{rossetto2010comparison}. Human-based assessments are not only inconsistent but also economically burdensome that requires specialized tools, technical expertise, and substantial travel and insurance costs \cite{drones5040106}. These limitations disproportionately affect low-income communities, which often lack access to the resources necessary for timely damage evaluation and recovery planning.

To address these disparities, this paper proposes an AI-driven framework that leverages drone-recorded video data, a resolution enhancement technique, and a Visual Language Model (VLM) to perform automated damage assessment. VLM is an emerging extension of Large Language Models (LLMs) that allow users to upload images or videos and receive intelligent, natural language analysis without requiring programming skills. With support for numerous languages, VLMs have the potential to democratize access to disaster assessment tools worldwide.

%MOVED THIS PARAGRAPH FROM THE RESULTS SECTION TO HERE. 
In this paper, we specifically focus on two types of commonly occurring natural disasters: tornadoes and earthquakes. We chose tornadoes as they are some of the most common natural disasters, with over 1,200 happening annually on average in the United States of America alone \cite{Chinchar_2023}. Due to the high number of occurrences of tornadoes as well as their suddenness to appear having a quick, effective, and cost-efficient way of assessing damages is of high importance. We also chose to focus on earthquake damages as every day there are approximately 55 earthquakes around the world, with an expected 15 of them being magnitude 7.0 and one of them being magnitude 8.0 \cite{USGS_2022}. Both tornadoes and earthquakes are major natural disasters that often happen with little to no warning and cause thousands of casualties and millions of dollars in damages.

The main contributions of our work are summarized as follows:

\begin{itemize}[nosep]
    \item We employ a super-resolution model to improve the quality of pre- and post-disaster video footage, thereby improving the accuracy of damage detection.
    \item We employ the emma3:27b visual language model to identify and describe structural damages in disaster-affected regions based on enhanced visual inputs. 
    \item We leverage the outputs of the VLM to evaluate the structural integrity of buildings and to guide and prioritize recovery and reconstruction efforts.
\end{itemize}

The remainder of this paper is organized as follows. Section~\ref{related_work} reviews related work in super-resolution imaging and AI-based damage detection. Section~\ref{methodology} presents the methodology for our proposed framework, including data collection, video enhancement, object detection, and VLM-based analysis. Section \ref{Datasets Descriptions} discusses the dataset used in this study.  Section~\ref{Results-KeyFindings} discusses the experimental results and key findings. Finally, Section~\ref{Conclusion-Future} offers concluding remarks and outlines future directions for extending this research.

\section{Related Work}
\label{related_work}

%\mamun{The subtitle of Super Resolution for Better Damage Detection and I-Based Damage Detection, what are the differences?}

% \subsection{Super Resolution for Better Damage Detection}
\noindent{{\bf{Super Resolution for Better Damage Detection}}.
Although AI Super Resolution itself does not identify any damages, nor does any classification, it gives either humans or AI damage detection software significantly higher quality imagery to make actual damage detection easier. Numerous studies have highlighted the advantages of applying AI-based super-resolution techniques for post-disaster damage detection. These include identifying subtle defects in concrete and building foundations \cite{https://doi.org/10.1155/2023/8850290, pawar2025deeplearningframeworkinfrastructure}, detecting both major and minor road surface damages \cite{Security_2024, drones6070171, s22239092}, and evaluating damage to critical infrastructure such as power lines \cite{zhou2024srgan}. These elements are crucial in disaster response and recovery, as they directly reflect the severity of the event and guide prioritization in mitigation and rebuilding efforts.

The detection of building foundation damage is particularly important, as even minor structural failures can escalate into catastrophic collapses, endangering nearby infrastructure and human lives. In \cite{https://doi.org/10.1155/2023/8850290}, the authors found that traditional damage detection algorithms often failed when applied to blurred, out-of-focus, or low-resolution images, even missing defects in otherwise high-quality visuals. To address this, they employed a Super-Resolution Generative Adversarial Network (SRGAN), which significantly improved the detection of previously unrecognized structural cracks. The authors in \cite{pawar2025deeplearningframeworkinfrastructure} further addressed challenges associated with computational inefficiencies and high false positive rates among existing super-resolution methods. They introduced an Efficient Sub-Pixel Convolutional Neural Network (ESPCNN), a lightweight yet high-performing super-resolution model, combined with a Convolutional Neural Network (CNN) tailored for damage classification. Their approach proved particularly effective on low-resolution inputs, confirming the potential of super-resolution-enhanced pipelines in resource-constrained post-disaster contexts. In the domain of road infrastructure, \cite{Security_2024} demonstrated that the enhancement of low-quality imagery using super-resolution techniques significantly improved the accuracy of road surface damage detection. However, they also noted a potential downside: if not carefully managed, image enhancement models can amplify noise and artifacts from the original data, potentially leading to misclassifications. Similar conclusions were reached in \cite{drones6070171}, where drone footage captured from 30 feet above ground was processed through a super-resolution pipeline. The authors reported notable improvements in image quality and detection accuracy, reaffirming the utility of AI-driven enhancement methods for aerial disaster assessment.

Both \cite{Security_2024} and \cite{drones6070171} conclude that super-resolution techniques are particularly effective for detecting fine-grained road damage from drone imagery. However, they caution that SRGAN-based models, while improving clarity, can introduce visual artifacts that compromise detection accuracy. In contrast, \cite{s22239092} compared multiple super-resolution approaches and found that while SRGAN provided the second-highest damage confidence scores, a customized model specifically designed for pavement analysis outperformed SRGAN, offering both greater precision and reliability.

% \subsection{AI-Based Damage Detection}

\noindent{{\bf{AI-Based Damage Detection}}. The application of artificial intelligence in post-disaster damage assessment has gained substantial attention in recent years due to its potential for rapid, scalable, and accurate analysis of large geographic areas. A particularly promising trend involves vision-based damage detection, where AI models analyze images and video footage to assess structural integrity without the need for manual inspection or physical access to affected sites.

Several approaches have been proposed in this area. For instance, the authors in \cite{SPENCER2025100203} demonstrated the use of unmanned aerial vehicle (UAV) imagery for structural damage detection, while \cite{buildings13051258} presented a taxonomy of damage types ranging from superficial to severe that AI models can effectively classify. Additionally, \cite{Ahn_Han_Park_Kim_Park_Cha_2025} explored cross-disaster scenarios, highlighting AI's ability to generalize damage detection across diverse natural disaster types. Complementarily, \cite{gholami2022deployment} employed high-resolution pre- and post-disaster satellite imagery to rapidly classify building-level damage, further validating the utility of image-based approaches in disaster response.

The authors in \cite{Umeike_2024} compared two deep learning methods for damage detection. The first method utilized You Only Look Once Version 11 (YOLOv11), offering real-time object detection and efficient localization of affected structures, with an achieved accuracy of 60.83\%. The second method employed a pretrained ResNet50 model, fine-tuned on the ImageNet dataset, which yielded a significantly higher accuracy of 90.28\%. These results illustrate the effectiveness of transfer learning and model selection in enhancing post-disaster assessment accuracy. However, this study primarily focused on tornado-related damage, whereas our proposed solution produces consistent and accurate results for both tornadoes and earthquakes. Additionally, by incorporating a visual language model (VLM), our approach is accessible to a broader, non-technical audience, enabling users to interpret and apply the results without requiring specialized expertise.
%NOTE; hopefully fixed with the last sentence I added. did use chat to make it sound more academic without adding any new context
%\mamun{However, this study only focuses on tornadoes whereas in our study we provide consistent results in both tornadoes and earthquakes...}

%\mamun{Provide a short paragraph about how this study is different than the other ones at the end of related works. }

%BELOW IS WHAT I ADDED
Unlike previous AI-based damage assessment methods that rely on single post-damage images, our approach leverages drone footage from both before and after a natural disaster to assess structural damage from multiple angles. This enables the detection of damage that may not be visible in a single view, such as destruction on the opposite side of a building. Furthermore, our method uniquely integrates a super-resolution model to enhance low-quality footage and a visual language model to generate interpretable, human-readable damage assessments. This combination not only improves analysis accuracy but also makes the results more accessible to non-technical users. 

\section{Methodology}
\label{methodology}

%ORIGINAL FIG
%\begin{figure*} [h]
%    \centering
%    \includegraphics[width=\linewidth]{figures/Paper_Figure.png}
%    \caption {High level overview of the proposed damage classification architecture}
%    \label{fig:enter-label}
%\end{figure*}

%NOTE: CHANGE "selction every nth frame" to "select frame" add whole and proper description to the paragraph describing, not the fig itself. 
\begin{figure} [!t]
    \centering
    \includegraphics[width=\linewidth]{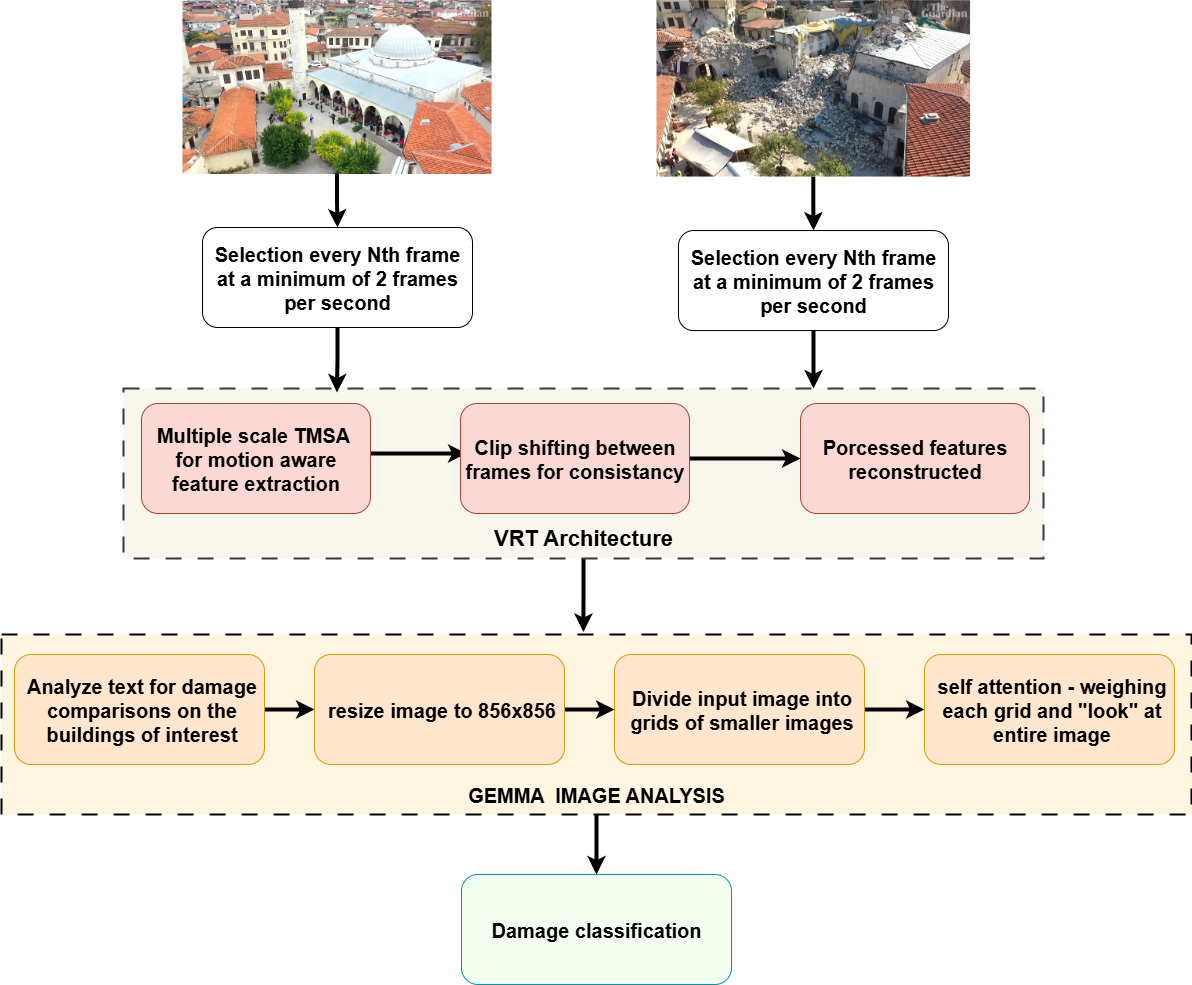}
    \caption {High level overview of the proposed damage classification architecture.}
    \label{fig:methodology}
\end{figure}

%\subsection{The Proposed Solution}

%\mamun{The methodology looks like introduction. Describe what we did here and use active tense. }

%In order to properly assess any damage that has occurred due to a natural disaster a before video of the area before the disaster must be recorded for accurate estimations. 

In this study, we use datasets consisting of both aerial drone footage and static satellite imagery that are captured before and after two distinct types of natural disasters: a tornado and an earthquake. Our methodology (Figure \ref{fig:methodology}) involves four key steps: (1) collecting relevant visual data; (2) selecting frames at a minimum rate of two per second to ensure adequate temporal coverage; (3) enhancing image quality using AI-based super-resolution techniques; and (4) analyzing the enhanced visuals using a visual-language model to extract semantic insights related to disaster impact.

To prepare the data for analysis, we extract every 10th frame from drone footage recorded by The Guardian, whose drone captures video at 25 frames per second. This results in approximately 2.5 frames per second, balancing temporal context with computational efficiency. We perform frame extraction using FFmpeg, an open-source video processing tool \cite{ffmpeg}. For static satellite imagery, we duplicate the original image within the dataset directory to simulate pseudo-frames, enabling compatibility with video-based processing pipelines. We then apply the VRT algorithm separately to the pre- and post-disaster frames. VRT is specifically designed for video enhancement by leveraging adjacent frames to improve image quality. Processing the pre- and post-disaster sequences independently ensures that damage present in post-disaster frames does not influence the enhancement of pre-disaster imagery.

For the image analysis and damage detection phase, we use Gemma3:27b \cite{gemma_2025}, a state-of-the-art open-source VLM architecture capable of interpreting multimodal data. We first prompt the model to analyze each structure in the pre-disaster frames in detail. Once this baseline is established, we provide the post-disaster frames and instruct the model to compare them, classifying each structure into one of four categories: (1) No/Slight Damage – Least Concern, (2) Moderate Damage – Moderate Concern, (3) Major Damage – High Concern, and (4) Totally Destroyed – Severe Concern. These categories are designed to guide disaster response efforts by indicating structural safety levels and prioritizing areas based on the potential risk to emergency and recovery personnel.

\section{Datasets Descriptions} \label{Datasets Descriptions}

Drone-based footage capturing both pre- and post-disaster conditions remains rare due to the historically limited demand for such data and the inherent unpredictability of most natural disasters. Given this limitation, we utilize the publicly available xBD dataset, which comprises thousands of annotated satellite images representing six major types of natural disasters \cite{gupta2019xbd}. For this study, we focus specifically on imagery from the $2013$ Moore tornado, an EF5 event that occurred in Moore, Oklahoma. These images, however, are static and lack temporal continuity, which restricts the effectiveness of AI-based super-resolution models that rely on adjacent frames to enhance image quality. Additionally, the single-view perspective limits the VLM's ability to assess structural damage comprehensively, as destruction visible from one angle may be obscured from another. The dataset includes $226$ paired pre- and post-disaster images, with each building labeled according to the damage scale adopted in this study. We select this subset to evaluate our proposed methodology on tornado damage a prevalent and high-impact disaster type in the United States while intentionally avoiding selective sampling of data known to perform well, thereby mitigating selection bias.

To further assess the applicability of our method to drone-specific data, we incorporate a video released by The Guardian, which captures aerial footage from before and after two severe earthquakes that struck Turkey in February 2023 \cite{GuardianNews_2023_Before}. These earthquakes, with magnitudes of 7.7 and 7.6, affected over 13 million people. The drone footage allows us to explore real-world conditions such as imperfect flight paths, motion instability, and multi-angle visual context factors often absent in still imagery datasets. This dynamic context enables both the super-resolution model and the VLM to more accurately identify damage that may be imperceptible in single-frame views. The original footage, downloaded at a resolution of 853×480 pixels, covers ten distinct disaster impacted regions, with each frame requiring approximately one megabyte of storage. We select this dataset prior to applying any analytical procedures to avoid bias and to ensure that earthquake damage characterized by widespread and unpredictable destruction across diverse geographic and socioeconomic contexts is adequately represented in our evaluation.

\section{Results and Key Findings}
\label{Results-KeyFindings}
\vspace{-3pt}
In this section, we present the results of applying our proposed methodology to two distinct natural disaster scenarios: tornadoes and earthquakes. We evaluate the approach using two complementary strategies. For tornado damage assessment, we compare the model's classification outputs against established ground truth labels provided in the xBD dataset. For earthquake damage, where such labels are not available, we assess structural impact using the Modified Mercalli Intensity (MMI) scale as a proxy for severity. The results indicate that our methodology demonstrates strong performance in both scenarios, achieving high classification accuracy and offering a reliable framework for rapid, initial post-disaster damage assessment.

\begin{figure}[!t]
   \centering
    \begin{subfigure}[b]{0.49\linewidth}
        \centering
        \includegraphics[width=\linewidth]{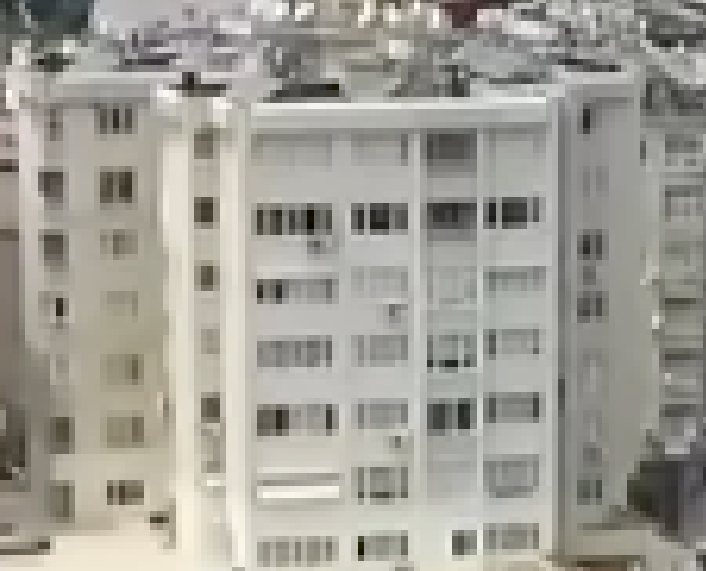}
        \caption{Building from drone footage before the earthquakes and before super resolution algorithm has been applied.}
        \label{fig:pre_quake_pre_super}
    \end{subfigure}
    \hfill
    \begin{subfigure}[b]{0.49\linewidth}
        \centering
        \includegraphics[width=\linewidth]{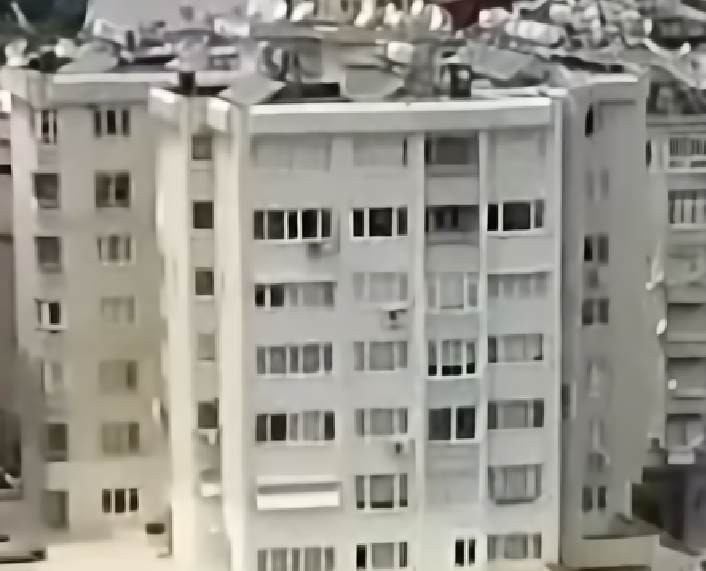}
        \caption{Building from drone footage before the earthquakes and after the super resolution algorithm has been applied.}
        \label{fig:post_quake_post_super}
    \end{subfigure}
    \caption{Comparison of drone footage before and after enhancement using VRT. The VRT-enhanced imagery reveals finer architectural details and enables the VLM to be more effective to interpret the buildings.}
    \label{fig:earthquake_comparison}
\end{figure}

\textbf{2013 Moore Tornado Damage Assessment and Comparison}. 
% Since our proposed solution operates on video input, we adapt it accordingly for testing with still images. Firstly, we no longer have to extract frames where we achieve at least 2 frames per second as we are working with individual images. To get around the limitations of VRT, the AI super resolution algorithm chosen, we must copy the individual frame a minimum of one time and place it in the same location as the original frame, resulting in a directory containing a minimum of 2 images. This is done to simulate multiple frames of a video that VRT requires for it's enhancement to work. Once we do this our proposed solution stays the same, giving the pre-disaster image to Gemma3:27b to spot the buildings and do a deep analysis on them then giving the post disaster image and having it rank the buildings into the categories used we have defined. 

To evaluate the effectiveness of our proposed method, we employ three key performance metrics: precision, recall, and F1-score (equation 1, 2 and 3). Precision quantifies the proportion of correct positive predictions, recall measures the proportion of actual positives that the model correctly identifies, and F1-score provides a harmonic mean of precision and recall. These metrics not only assess classification accuracy but also align with the evaluation standards used in the benchmark dataset adopted for this study \cite{gupta2019xbd}. Compared to the ground truth provided by the xBD dataset, we achieve the results presented in Table~\ref{tab:performance comparison between proposed solution and baseline}.

\begin{align}
%\text{Accuracy} &= \frac{TP + TN}{TP + TN + FP + FN} \\
\text{Precision} &= \frac{TP}{TP + FP} \\
\text{Recall} &= \frac{TP}{TP + FN} \\
\text{F1} &= 2 * \frac{(precision*recall)}{precision + recall} 
\end{align}

When comparing the results presented in Table~\ref{tab:performance comparison between proposed solution and baseline}, we observe that our proposed solution significantly outperforms the baseline in terms of F1 scores and precision. Although our model demonstrates weaker precision in the No Damage and Major Damage classes, this discrepancy arises from the subtle visual differences between major and minor damage, which often lead to misclassification, as noted in prior research \cite{gupta2019xbd}. This limitation is further compounded by the fact that our evaluation relies on only a single pre-disaster and post-disaster image per building. In contrast, the intended deployment of our method assumes access to multiple images of the same building from various angles and distances. Such a configuration would provide the visual language model with richer contextual information, thereby improving its ability to distinguish between damage categories particularly between no damage and major damage.

We omit a comparison of accuracy scores, as the authors of the baseline dataset highlight that this metric can be misleading. Given the high proportion of unaffected buildings in disaster zones, a model that classifies all structures as ``No Damage” could still attain an artificially inflated accuracy of 75\%.

\begin{table}[!t]
    \centering
    \resizebox{\columnwidth}{!}{
    \begin{tabular}{r|rrr|rrr}
    \toprule
      Damage Type  &  \multicolumn{3} {c|} {Baseline Performance Metrics} & \multicolumn{3}{c} {Proposed Solution Performance Metrics} \\ 
      & Precision & Recall & F1 Score & Precision & Recall & F1 Score \\ \midrule
      No Damage & 0.8770 & 0.5330 & 0.6631 & 0.737 & \textbf{0.875} & \textbf{0.800} \\
      Minor Damage & 0.1971 & 0.1128 & 0.1435 & \textbf{0.750} & \textbf{0.536} & \textbf{0.626}\\
      Major Damage & 0.7259 & 0.0047 & 0.0094 & 0.571 & \textbf{0.667} & \textbf{0.615}\\
      Destroyed & 0.5050 & 0.4321 & 0.4657 & \textbf{0.893} & \textbf{0.895} & \textbf{0.894} \\ \midrule
      Overall Accuracy & & \textbf{N/A} & & &\textbf{84.5\%} &
     \\ \bottomrule
    \end{tabular}
    }
    \caption{Performance comparison of the proposed solution and baseline by damage classes}
    \label{tab:performance comparison between proposed solution and baseline}
\end{table}

% \subsection{2023 Turkey Earthquakes Damage Assessment and Comparison}
\noindent{\textbf{2023 Turkey Earthquakes Damage Assessment and Comparison}}.
% In this study, we compared the proposed solution's effectiveness with expert human damage assessments and footage analysis based off the MMI scale, a relative scale ranking damages from a disaster from opinions of human experts. 
In this study, we compare the effectiveness of the proposed solution with expert human damage assessments and visual analyses based on the MMI scale, a qualitative scale that ranks disaster-induced damage according to expert opinion and observational data.

\subsubsection{Footage Analysis From Our Proposed Solution}
To reduce the computational resources required by both the super-resolution algorithm and the visual language model (VLM), we extract and retain every tenth frame from the original video footage. Given that the original footage operates at 25 frames per second (fps), this approach yields an average of 2.5 frames per second. This significantly decreases the memory footprint and processing demand while preserving sufficient temporal context to enable high-quality visual analysis. Each saved frame is subsequently processed through the selected super-resolution algorithm, resulting in an upscaled resolution of 3412×1920 pixels. This represents a 16-fold increase in pixel count, substantially enhancing image clarity and revealing fine-grained structural damage that might otherwise remain undetected. With the upscaled frames prepared, we employ Gemma 3:27b, a 27-billion-parameter large language model developed by Google—for scene understanding and damage assessment. We first present the model with a pre-disaster image and prompt it to perform a detailed analysis of the visible structures, with a focus on identifying buildings. This preliminary step establishes a reference for comparison in cases where structures may be completely destroyed in the post-disaster scenario.

In its analysis of the pre-disaster image, Gemma 3:27b accurately detects several key features: the roundabout and its central monument, a modern building behind the roundabout (suggested to be a government facility based on architectural style and location), a distinctive semi-circular structure in the bottom-right quadrant of the image, several high-rise and multi-story buildings in the background, and a group of commercial buildings lining the road extending to the right. These identifications serve as baseline markers for post-disaster comparison. Next, we present the post-disaster image to Gemma 3:27b and instruct it to assess the condition of each previously identified structure. The model categorizes the damage into four levels of severity, corresponding to levels of concern for emergency response:
\begin{enumerate}
    \item No/Slight Damage - Least Concern
    \item Moderate Damage - Moderate Concern
    \item Major Damage - High Concern
    \item Totally Destroyed - Severe Concern
\end{enumerate}

Gemma 3:27b correctly infers that the observed damage likely results from a major earthquake. It classifies buildings in the far background (particularly in the north and east) as having sustained no or slight damage—suggesting these were distant from the epicenter and thus experienced minimal impact. Only 5\% to 10\% of buildings fall into this category, representing areas of least concern for recovery efforts. Approximately 30\% to 40\% of buildings are categorized as moderately damaged. These include background structures displaying visible cracking, leaning facades, and partial collapses—particularly on upper floors—while retaining overall structural integrity. The model concludes that the majority of midground and background buildings fall within this category, indicating a moderate level of concern for rescue operations. A significant portion approximately 40\% to 50\% of the buildings are classified as severely damaged. These structures, particularly concentrated in the midground and foreground, exhibit substantial structural compromise, including pronounced cracks, visible leaning, and signs of imminent collapse. Notably, the building behind the roundabout is placed in this category. The model emphasizes that these structures pose a high risk to both recovery personnel and the surrounding environment.

Finally, Gemma 3:27b identifies several buildings as totally destroyed, warranting the highest level of concern. These structures are of critical interest for search-and-rescue operations and may involve hazardous materials or unstable debris. The model classifies the semi-circular building to the right, as well as multiple structures behind the roundabout, within this category. Gemma 3:27b also notes potential limitations in its analysis particularly regarding structures located near the edges of the image or in the far background. Due to the limited vantage point and single flyover perspective of the drone footage, some damage may be underrepresented or misclassified.

\begin{figure}[h!]
  \centering
  \includegraphics[width=0.49\textwidth]{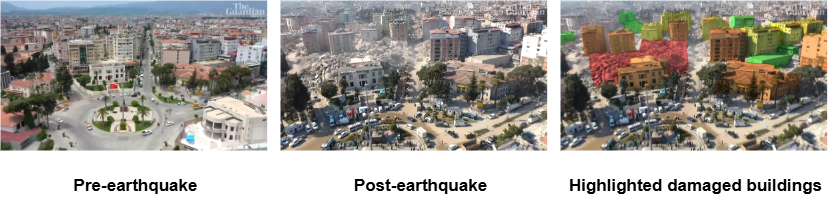}
  \caption{Buildings manually annotated based on the Gemma-3:27b model's textual description of structural damage. Color coding reflects estimated damage levels: green indicates no or slight damage, yellow indicates moderate damage, brown indicates severe damage, and red indicates total destruction.}
  \label{fig:myfigure}
\end{figure}

\subsubsection{Comparing Our Proposed Solution Effectiveness on Earthquakes with Human Experts}

\begin{figure*}[!t]
    \centering
    \begin{subfigure}[b]{0.49\textwidth}
        \centering
        \includegraphics[width=\linewidth]{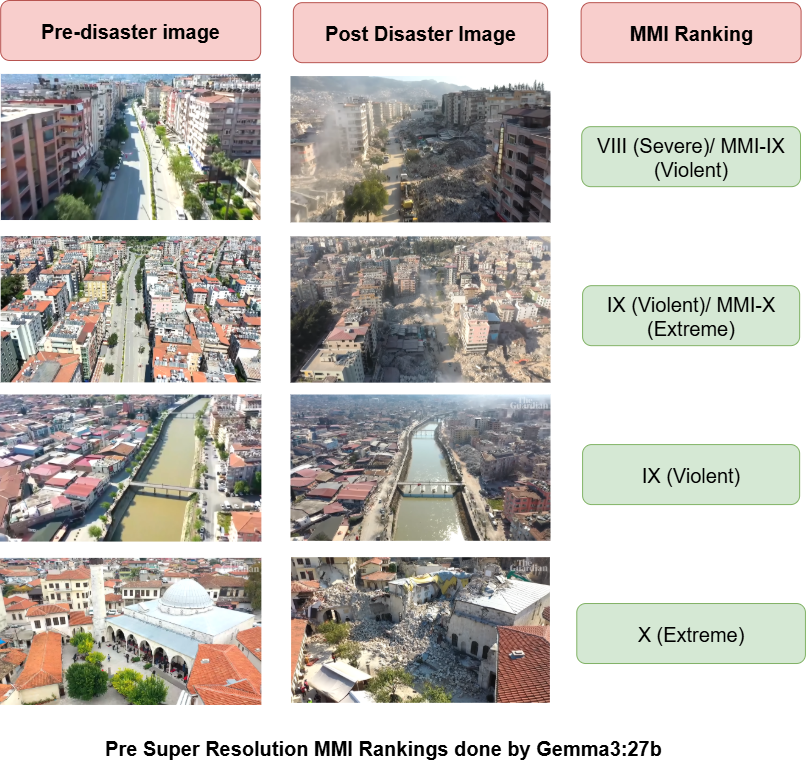}
        \caption{Example frames given to Gemma3:27b and what it classified each place on the MMI scale before the super resolution algorithm has been applied}
        \label{fig:MMI Rankings pre-super resolution}
    \end{subfigure}
    \hfill % This command creates a horizontal space between the two images
    \begin{subfigure}[b]{0.49\textwidth}
        \centering
        \includegraphics[width=\linewidth]{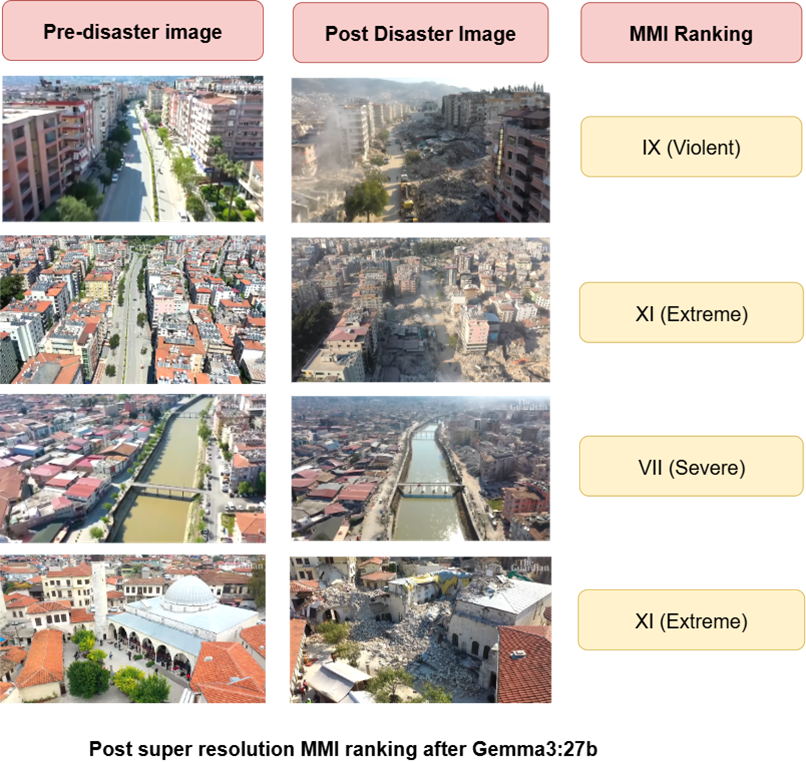}
        \caption{Example frames given to Gemma3:27b and what it classified each place on the MMI scale after the super resolution algorithm has been applied}
        \label{fig:MMI Rankings post-super resolution}
    \end{subfigure}
    \caption{Comparison of damage assessments based on the MMI scale using identical frames from earthquake-affected regions, before and after application of a super-resolution algorithm. The enhanced resolution led to observable changes in MMI scores, indicating a potential impact of image quality on perceived damage severity. %{\color{red} 3 Figures of severe, extreme and violent will be fine.}
    }
    \label{fig:earthquake_comparison}
\end{figure*}

According to the Kandilli Observatory, the earthquakes reached the MMI XII (Extreme) category on the MMI scale at the epicenter, the highest categorization of an earthquake affects at a location. Other locations farther away from the epicenter were classified from MMI-VI (Strong) to MMI-XI (Extreme) \cite{PreliminaryAnalysis}. According to Gemma3:27b after being shown the frames of a video before and after the natural disaster of 10 locations in turkey it classified the damages in the locations shown in figure 4 from MMI-VIII (Severe) to MMI-XI (Extreme), with an overall classification of MMI-XI (Extreme). Since the MMI scale is based on arbitrary damage measurements decided by human experts, slight differences are to be expected not only with the VLM assessment, but also between human experts themselves \cite{EarthquakeHazardsProgram}. The difference is also described the fact that the MMI scale is different between areas affected by a natural disaster, and since we analyzed several different areas away from the epicenter of the earthquake we are expected to get different results. 
This proposed solution has shown that it's able to accurately identify buildings pre-disaster, save that to memory, able to correlate buildings from the pre-disaster frames to buildings post-disaster frames, use key locations and landmarks to determine positions of buildings, able to rank buildings into 1 of 4 categories to help with recovery efforts post disaster by knowing where is potentially hazardous and where might be safe, and finally showing itself to be highly accurate in terms of ranking damages using a scale and comparing it's findings to that made by human experts resulting in very similar results in an arbitrary scale.

\subsection{Visual Language Model Damage Detection Without Super Resolution}
In order to test the super resolution algorithm effectiveness on the VLM damage detection we use the earthquake footage provided by The Guardian and ask the VLM to do the same damage detection with the same prompts. In order to do this comparison we have sent the same prompt multiple times to Gemma3:27b to gauge it's MMI rankings for each footage, and put the results in a table for what the VLM has stated for the original footage and the super resolution footage, allowing direct comparisons in exactly how the super resolution algorithm has changed the rankings for each of the 10 disaster before and after frames. We ensured that new sessions were created between each MMI ranking to prevent bias of remembering what it has previously ranked the images in other sessions. We also send both the pre and post super resolution frames at the same time to Gemma3:27b and ask it to direct compare the different results and how the super resolution algorithm changed it's MMI rankings and why, this is done in order to know specifically how the different resolutions have impacted the results according to Gemma3:27b itself. 
Through this comparison we can see that the super resolution image resulted in higher MMI rankings for most of the locations captured by the drone footage, being closer to the max MMI ranking given to the overall earthquakes. The VLM specifically stated that due to the increase in resolution it was able to better detect and no longer rely on as many assumptions as it had to do with the lower quality original footage. Through this increased resolution the VLM was more confident in the higher damage assessments that matches closer to the professional assessment rather than having to make assumptions on buildings in the background being more intact then they actually were resulting in a lower overall damage assessment.

%mention what classifications mean/how lead stuff to it
%in figure make clear the changes in ranking clear by bolding or changing colour

%line chart showing different stuff
%TALK WHY OUR MODEL DOES DIFFERENT (ranking damages), time it takes, etc. so that even if we get worse results why it's still an important research that others know about. 

\vspace{-4pt}
\section{Conclusions and Future Work}
\label{Conclusion-Future}
With the recent advances of artificial intelligence, specifically advances in super-resolution and visual language models, we have shown that they can play a pivotal key role in recovery efforts by classifying damages, showing critical areas to start recovery efforts to prevent more damages and casualties, as well as being able to survey a large affected area faster and cheaper than human experts. 
The use of drones for video footage has allowed us to survey a large area at once, quickly and in areas that might be inaccessible to typical damage assessments due to blockages. Once the drone footage has been captured, the use of Video Restoration Transformer (VRT) as a super resolution algorithm has shown that even from low quality footage, either from inexpensive drones or footage taken from a high altitude to survey a larger area at once, we are still able to gather important information about affected buildings and the damages done without sacrificing accuracy and costs on more expensive drones. From the upscaled footage the visual language model used was able to not only identify the damaged buildings but also was able to successfully classify them into one of four categories, from no/slight damaged to totally destroyed, providing important information to disaster recovery experts. The VLM has also shown to be very generalist in it's ability to classify damages, successfully classifying damages from a multitude of different disasters such as earthquakes and tornadoes, allowing a simple solution to be trained on instead of workers having to be trained on several different analysis methods for different disasters.

Future work should focus on different natural disasters to test if the proposed solution isn't able to do accurate assessments on certain disasters. Future work should also test different Visual Language Models as well as different AI Super Resolution algorithms to study if different VLMs and super resolution algorithms are better or worse at analyzing damages caused by a natural disaster then others. Finally future research should use the proposed methodology to capture unique pre and post natural disaster footage in different regions in order to test the proposed solution's ability to understand different damages from different standards of structures. 

\section*{Acknowledgements} \label{Acknowledgements}
The authors acknowledge the use of generative AI tools to assist in improving the fluency and grammar of the manuscript. No content was generated beyond these editorial refinements. The authors thank Google team for Gemma 3:27b.

% \newpage

\bibliographystyle{ieeetr}
\bibliography{references}

\end{document}